\documentclass{article}

\usepackage{PRIMEarxiv}
\usepackage[numbers,sort&compress,sectionbib]{natbib}
\usepackage[utf8]{inputenc} 
\usepackage[T1]{fontenc}    
\usepackage{hyperref}       
\usepackage{url}            
\usepackage{booktabs}       
\usepackage{amsfonts}       
\usepackage{nicefrac}       
\usepackage{microtype}      
\usepackage{lipsum}
\usepackage{fancyhdr}       
\usepackage{graphicx}       
\graphicspath{{media/}}     
\usepackage[table]{xcolor}
\usepackage{colortbl} 
\title{Explainable vision transformer enabled convolutional neural network for plant disease identification: PlantXViT
}

\author{
	Poornima Singh Thakur, Pritee Khanna, Tanuja Sheorey, Aparajita Ojha\\
	PDPM Indian Institute of Information Technology, Design and Manufacturing \\
	Jabalpur, India 482005\\
	\texttt{\{poornima, pkhanna, tanush, aojha\}@iiitdmj.ac.in} \\
}

\begin{document}
	\maketitle

	\begin{abstract}
		Plant diseases are the primary cause of crop losses globally, with an impact on the world economy. To deal with these issues, smart agriculture solutions are evolving that combine the Internet of Things and machine learning for early disease detection and control. Many such systems use vision-based machine learning methods for real-time disease detection and diagnosis. With the advancement in deep learning techniques, new methods have emerged that employ convolutional neural networks for plant disease detection and identification. Another trend in vision-based deep learning is the use of vision transformers, which have proved to be powerful models for classification and other problems. However, vision transformers have rarely been investigated for plant pathology applications. In this study, a Vision Transformer enabled Convolutional Neural Network model called "PlantXViT" is proposed for plant disease identification. The proposed model combines the capabilities of traditional convolutional neural networks with the Vision Transformers to efficiently identify a large number of plant diseases for several crops. The proposed model has a lightweight structure with only 0.8 million trainable parameters, which makes it suitable for IoT-based smart agriculture services. The performance of PlantXViT is evaluated on five publicly available datasets. The proposed PlantXViT network performs better than five state-of-the-art methods on all five datasets. The average accuracy for recognising plant diseases is shown to exceed 93.55\%, 92.59\%, and 98.33\% on Apple, Maize, and Rice datasets, respectively, even under challenging background conditions. The efficiency in terms of explainability of the proposed model is evaluated using gradient-weighted class activation maps  and Local Interpretable Model Agnostic Explanation.
	\end{abstract}

	
	\label{introduction}
	Human population will surpass 10 billion in the next 30 years, which indicates a higher food demand \cite{WPOP} in coming decades. To meet the future requirements and for the sustainable agriculture, methods to prevent the crops from pests and diseases are of utmost importance, as plant diseases have a considerable impact on crop production quality and quantity worldwide. Plant diseases alone are responsible for 20–40\% of crop yield losses, \cite{FAOnew} which affects the agriculture industry on a large scale. To deal with these issues, smart agriculture solutions are being explored worldwide that combine the Internet of Things (IoT) and machine learning (ML) based methods for early disease detection and control. This has led to significant advancements in the area of vision-based plant disease detection methods for real-time disease detection and diagnosis. 
	
	Several ML approaches have been suggested by researchers over the years. Amongst them, support vector machines \cite{zhang2016cucumber,sun2019slicsvm,kumar2020plant,hou2021recognition,hamdani2021detection}, artificial neural network (ANN) \cite{ramesh2020recognition, hamdani2021detection}, Naive Bayes \cite{johannes2017automatic,abdu2020automatic,hamdani2021detection}, k-means clustering \cite{johannes2017automatic,ramesh2020recognition}, and simple linear iterative clustering \cite{johannes2017automatic,sun2019slicsvm,hou2021recognition} are some of the most extensively used methods. In recent years, the focus is shifted towards deep learning (DL) algorithms due to the availability of large amounts of data, computing power, and efficient training approaches. The powerful feature learning capabilities of convolutional neural network (CNN) architectures have produced prominent results in plant disease detection. Apart from standard architectures such as AlexNet, GoogleNet, VGG16, ResNet with transfer learning approaches \cite{mohanty2016using,barbedo2018impact}, customized CNN architectures have also been introduced for plant disease detection tasks \cite{huang2020detection,yadav2021identification}. More recently, CNN models with attention mechanisms have been proposed that exhibit excellent performance in plant disease detection \cite{karthik2020attention,chen2020attention,chen2020identification,chen2021identification,chen2021identifying,zhao2022ric}. 
	
	Despite these developments, plant disease detection remains a challenge due to the variety of disease types for different crops, evolution of new diseases, and unavailability of in-field data for most of the crops. While DL models are data-hungry, lightweight CNN models do not generalize well for all types of crops. There are two main challenges that keep the research community busy; (1) interpretability of decisions made by ML systems to understand when the system may fail and (2) developing efficient lightweight CNN models that can generalize well for a large variety of plants and their disease types. 
	
	The concept of transformers in natural language processing \cite {vaswani2017attention} has opened new vistas in image processing and compute vision as its analogue Vision Transformers (ViT) has been recently introduced by \citet{dosovitskiy2020image} and has achieved exceptional classification performance with a significantly lower memory footprint on benchmark datasets of ImageNet, CIFAR-10, CIFAR-100, Oxford-IIIT Pets, Oxford Flowers-102, and VTAB. But the main issue with the ViT model is that it cannot converge well on small datasets when compared with CNN models. The reason could be that the ViT mainly focuses on the extraction of long-distance feature dependencies but lacks in efficiently capturing local features of the images \cite{neyshabur2020towards}. But the local feature extraction capability of CNN combined with the powerful self-attention modules of ViT may help in simultaneously extracting local and global features from images. Further, ViT networks alongwith with CNN can help in enhancing the explainability of plant disease detection models. Keeping this in view, a lightweight, explainable ViT-based plant disease detection model is introduced in the present paper that combines the feature extraction capabilities of CNN and ViT. The model consists of the initial two blocks of the pre-trained VGG16 network, followed by an inception block and four stacks of transformer encoders. The proposed model not only outperforms some of the recently introduced DL models \cite{karthik2020attention,chen2020attention,chen2021identification,chen2021identifying,zhao2022ric} on five publicly available datasets, but  also improves the explainability of its prediction. The main contributions of the present work are summarised as follows. 
	
	\begin{itemize}
		\item[$\bullet$] A ViT enabled CNN model, PlantXViT is proposed for plant disease identification that significantly improves the classification performance over a broad range of crop varieties and their diseases. 
		
		\item[$\bullet$] PlantXViT exhibits better explainability of its prediction through an analysis of gradient-weighted class activation maps and local interpretable model-agnostic explanations. 
		
		\item[$\bullet$] The model is lightweight with only 850,500 trainable parameters, that makes it suitable for smart agriculture devices.
		
		\item[$\bullet$] The proposed model outperforms some of the recently introduced deep CNN models, as demonstrated through extensive experiments on five public datasets with images under different capturing conditions. 
		
	\end{itemize}      
	
	The rest of the paper is organized as follows: Section \ref{related} provides a brief overview of the existing schemes for crop disease identification. Section \ref{proposed} is devoted to the proposed method. Section \ref{result} deals with experiments and results. Here we present the performance of the proposed model on five different datasets. Section \ref{conclusion} provides the concluding remarks and the future scope of the work.
	
	\section{Related works}
	\label{related}
	
	With the impressive performance of CNN in computer vision, researchers are increasingly interested in developing DL models for automatic plant disease detection and identification. A brief overview of the state-of-the-art CNN models for plant disease detection is presented in this section. Table  \ref{tab1} shows these techniques using various CNN architectures.
	
	The initial work on plant disease detection using CNN was carried out by \citet{mohanty2016using} on a large-scale dataset, by comparing the performances of AlexNet and GoogleNet models. They identified that the GoogleNet model with a transfer learning approach achieved a remarkable accuracy of 99.35\% on PlantVillage dataset with 54,305 images in 38 classes. In another work by \citet{barbedo2018impact} analyzed the impact of dataset size and diversity on plant disease detection using a transfer learning approach with the base model as the GoogleNet model. He showed that the model achieved the highest accuracy of 87\% on 1383 images in 56 classes with no background.He concluded that the limitation of the datasets in the area was the main bottleneck in practical deployment of CNN models. \citet{too2019comparative} studied the performance of standard architectures VGG16, Inception v4, ResNet with 50, 101, and 152 layers, and DenseNet with 121 layers on the same dataset using the fine tuning technique. As per their results, the highest accuracy of 99.75\% was achieved by the DenseNet121 model among all.
	
	In a work by \citet{chen2020using}, the initial two blocks of pre-trained VGG16 and two Inception v3 blocks were combined to build a plant disease detection model. They evaluated the model's performance on three different datasets. Their model achieved 84.25\%, 92\% and 80.38\% accuracy on Maize species from PlantVillage, Maize dataset, and their own Maize dataset respectively. \citet{thakur2022vgg} also developed a CNN model with the initial two pre-trained layers of VGG16 and Inception v7 blocks. They evaluated the performance of their model on five publicly available datasets and reported 99.16\%, 93.66\%, 94.24\%, 91.36\% and 96.67\% accuracy scores on PlantVillage, Embrapa, Apple, Maize and Maize datasets respectively. 
	
	\begin{table}[!ht]
		\centering
		\caption{An overview of related works on crop disease identification}
		\label{tab1}
		\scalebox{0.86}{
			\begin{tabular}{p{3cm}p{3.5cm}p{3.5cm}p{1cm}p{1cm}p{1cm}p{1cm}p{1.5cm}}
				\hline
				\rowcolor{gray!50}
				\textbf{Author}   & \textbf{Technique} & \textbf{Dataset} & \textbf{Acc \%}  & \textbf{Pre \%}  & \textbf{Re \%}  & \textbf{F1 \%}  & \textbf{Other}  \\
				\hline		
				\rowcolor{gray!25}
				\citet{mohanty2016using} & AlexNet and GoogleNet & 54,305 leaf images in 38 classes from PlantVillage dataset &  - & - & - & - & -  \\
				
				\rowcolor{gray!25} 
				& GoogleNet &  & 99.35 & 99.35 & 99.35 & 99.34 & -\\
				
				\citet{barbedo2018impact} & GoogleNet & 1383 images in 56 classes  & 87 & - & - & - & - \\
				
				\rowcolor{gray!25}
				\citet{too2019comparative} &  VGG16, Inception v4, ResNet with 50, 101 and 152 layers and DenseNet with 121 layers & 54,305 leaf images in 38 classes from PlantVillage dataset &  - & - & - & - & - \\
				\rowcolor{gray!25}
				& DenseNet121 & & 99.75 & - & - & - & Loss: 0.0159 \\
				
				\citet{chen2020using} & VGG19, Incption v3  & 466 maize leaf images in 4 classes  & 80.38 & - & 60.76 & - & Specificity 86.92 \\
				&  &  500 rice leaf images in 5 classes  & 92 & - & 80 & - & Specificity 95 \\
				
				\rowcolor{gray!25}
				\citet{karthik2020attention} & Residual CNN, attention mechanism & 95,999 tomato leaf images in 10 classes from PlantVillage dataset  & 98 & - & - & - & - \\
				
				\citet{zeng2020crop} &  Self-Attention CNN & 9214 leaf images in 6 classes from AES-CD9214 & 95.33 & - & - & - & - \\
				&  &  988 leaf images from MK-D2 dataset  & 98 & - & - & - & - \\
				
				\rowcolor{gray!25}
				\citet{chen2020attention} & Mobile-DANet, DenseNet, transition layer, depthwise separable convolution, attention module &  3852 maize leaf images 4 classes from PlantVillage dataset  & 98.5 & 97 & 97 & 97 & Specificity 99 \\
				\rowcolor{gray!25}
				& & 133 maize leaf images in 8 classes & 95.86 & 83.45 & 83.45 & 83.45 & Specificity 97.63 \\
				
				\citet{chen2021identification}  & MobileNet-V2, attention mechanism & 1045 leaf images in 10 classes from PlantVillage dataset & 99.67 & 98.37 &  98.37 & 99.81  & Specificity 99.81  \\
				&  &  1107 rice leaf images in 12 classes & 98.48  & 90.56  & 90.56  & 90.56  & Specificity 99.17 \\
				
				\rowcolor{gray!25}
				\citet{chen2021identification1} & MobileNet, SE & 1645 images in 11 classes from PlantVillage dataset & 99.78 & - &  98.83 & - & Specificity 99.88 \\
				\rowcolor{gray!25}
				& & 444 images in 25 classes from rice dataset & 99.33 & - &  87.87 & - & Specificity 99.67 \\
				
				\citet{chen2021identifying} & MobileNet v2, Depthwise separable convolution, channel and spatial attention module & 1045 leaf images in 10 classes from PlantVillage dataset & 99.71 & - & 98.56 & 98.56 & - \\ 
				& & 405 images in 20 classes & 99.13 & - & 91.37 & 91.37 & -\\
				
				\rowcolor{gray!25}
				\citet{thakur2022vgg} & VGG16, Inception v7 & 54,305 leaf images in 38 classes from PlantVillage dataset & 99.16 & - & - & - & Loss: 0.05 \\
				\rowcolor{gray!25}
				&  & 560 Rice leaf images in 4 classes & 96.67 & - & - & - & Loss: 0.05 \\

				\citet{zhao2022ric} & Inception, residual, modified channelwise attention module & 38,466 images of corn, potato and tomato from PlantVillage dataset in 17 categories & 99.55  & - & - & - & Loss 0.0175 \\
				
				\rowcolor{gray!25}
				\citet{lu2022hybrid} & GhostNet, ViT &  GLDP12k dataset with 12,615 vine leaf images in 11 classes & 98.14  & - & - & - & - \\
				\hline	
			\end{tabular}
		}
	\end{table}
	
	Recent trends have been on the use of attention mechanisms to boost the performance of plant disease detection models. In the attention mechanism, high priority is assigned to pixel locations with relevant information. Researchers have effectively used the attention mechanism to improve the classification performance of CNNs. \citet{karthik2020attention} have developed a residual CNN with an attention mechanism for tomato disease detection. Their model attains 98\% accuracy on a dataset containing 95,999 tomato leaf images labelled into 10 classes. The dataset is created using the images from PlantVillage dataset and applying augmentation on selected images. \citet{zeng2020crop} have also experimented with residual CNN and self-attention modules. Their model is shown to achieve 98.0\% and 95.33\% accuracy on the MK-D2 and AES-CD9214 datasets, respectively. 
	
	\citet{chen2020attention} have employed spatial and channel-wise attention modules with depth-wise separable convolution in DenseNet. Their model performs well on maize species of PlantVillage dataset, with 98.50\%. On their own collected Maize dataset, 95.86\% accuracy is reported by the authors. In another work by \citet{chen2021identification}, spatial and channel-wise attention mechanisms are applied to the MobileNet model. Their model also performs well, with 98.48\% accuracy on Rice dataset. In the continuation with this work, \citet{chen2021identification1} have also applied the squeeze-and-excitation (SE) block on MobileNet v2 to achieve 99.33\% accuracy on a custom Rice dataset. In yet another paper, the MobileNet v2 model is used with depthwise separable convolution and spatial and channel-wise attention, and this model also shows excellent performance with 99.71\% accuracy on a subset of PlantVillage \cite{chen2021identifying}. In another custom dataset containing disease images from paddy, corn, and cucumber plants, their model achieves 99.13\% accuracy \cite{chen2021identifying}.  In a more recent work by \citet{zhao2022ric}, a CNN model with inception modules and residual blocks has been developed with the attention mechanism. In their model, a modified convolutional block attention module is used that helps in achieving 99.55\% accuracy on corn, tomato, and potato datasets. With the introduction of ViT as a powerful image classification model, \citet{lu2022hybrid} have devised a plant disease detection model using the GhostNet and ViT blocks. The model has been evaluated on the GLDP12k dataset with 12,615 vine leaf images in 11 classes and achieves 98.14\% accuracy. 
	
	A review of existing works indicates that CNN models with attention mechanisms have demonstrated higher accuracy for specific types of crops and their diseases. However, these models have not been analysed for the cases when they may fail. The confusion matrix only gives an idea of the number of miss-classifications for a particular class. But the model's behavior on miss-classified samples has not been explored enough. In other words, despite the high accuracy achieved by several recent models, the interpretability of those models has not been explored. With advances in explainable AI, it is crucial to develop methods that not only perform well on a variety of plant diseases, but also aid scientists in analyzing the reasons for high accuracy and possible failures in certain cases. In the present paper, a ViT-enabled CNN model is proposed that significantly improves the disease classification performance on a wide variety of plant diseases. Furthermore, prediction results are shown to be explainable, and are compared with those of existing CNN models. 
	
	\section{ Explainable vision transformer enabled convolutional neural network: PlantXViT}
	\label{proposed}
	
	This section is devoted to the proposed model PlantXViT that uses ViT for plant disease detection and identification. The model consists of a CNN followed by the ViT. For the sake of completeness, a brief introduction to the ViT is presented first.  
	
	\subsection{Vision Transformer}
	
\begin{figure}[!h]
		\centering
		\includegraphics[width=0.96\linewidth]{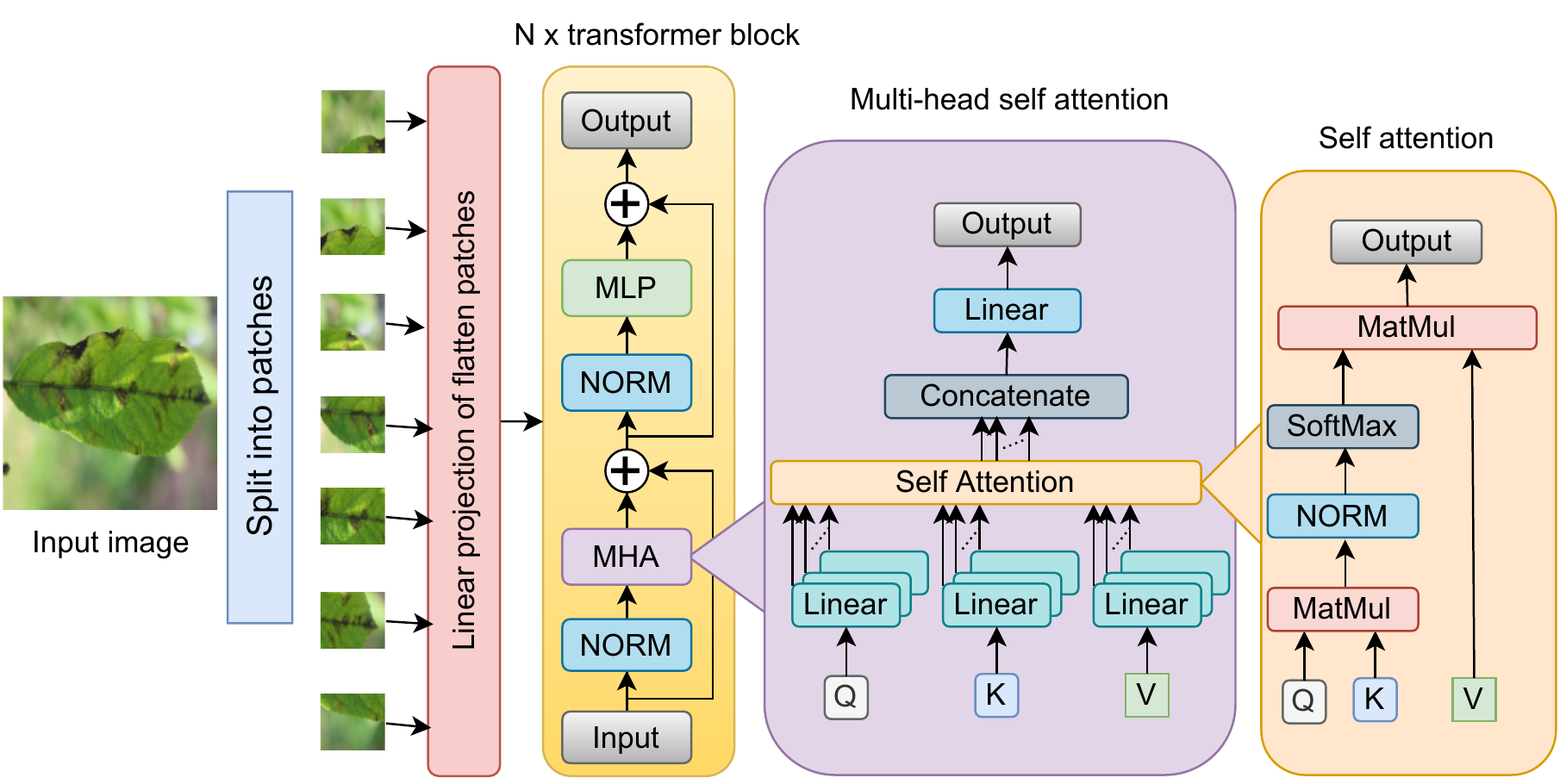}
		\caption{ ViT block with multi-head self-attention block and self-attention }
		\label{fig1}
	\end{figure}

	With the widespread success of transformer networks in natural language processing problems \cite{vaswani2017attention}, \citet{dosovitskiy2020image} developed the ViT model based on the architecture of the original transformer.  The ViT is composed of self attention blocks and multilayer perceptron (MLP) networks with a linear projection and positional embedding mechanism for an input image. The organization of a typical ViT is presented in Fig. \ref{fig1}. As presented, the input image is divided into fixed size non-overlapping patches. Further, patches are flattened and positional embedding is performed with linear projection. The positional embedding is basically used to retain the positional information of patches with respect to the original image. The output vector is then passed to a stack of $N$ number of transformer blocks. The main components of a typical transformer block are multi-head self-attention (MHA) and MLP. Each one is preceded by a normalisation layer and residual connection at the end. MHA includes self-attention, which is applied to each patch individually. In MHA, the input vector is transformed into three separate vectors: query (Q), key (K), and value (V). They are computed as $Q = XW_Q$, $K=XW_K$, and $V=VW_Q$; where $W_Q$, $W_K$, and $W_V$ are the weight matrices. A dot-product of Q and K is taken to generate a score matrix based on the saliency of the embedded patch. Then, the SoftMax activation function is applied to the score matrix. Further, the output is multiplied into $V$ to generate the self-attention result as shown in Eq. \ref{eq1} where $d_{k}$ represents the dimension of the vector $K$. 
	
	\begin{equation}
		\label{eq1}
		SA(Q,K, V) =  SoftMax(\frac{QK^{T}}{\sqrt{d_{k}}}) * V
	\end{equation}
	
	Finally, self-attention matrices are combined and passed onto a linear layer followed by a regression head. Self-attention enables the selection of relevant semantic features at image locations for classification. There can be any number of self-attentions present in the transformer encoder, known as MHA. Output of the MHA block can be calculated using Eq. \ref{eq2}. MLP is stacked in the transformer block after the MHA layer. MLP includes ANN layers with a GeLU activation function. The GELU activation is calculated by multiplying the input by its Bernoulli distribution. It has skip connections from the output of MHA, as presented in Fig. \ref{fig1}. The output of the transformer block can be calculated using Eq. \ref{eq3}.
	
	\begin{equation}
		\label{eq2}
		MHA_{out} = MHA(NORM(x_{in})) + x_{in}
	\end{equation}
	
	Where $x_{in}$ is the input to transformer block $NORM$ is the normalization layer, $MHA$ is multi-head self-attention, and $MHA_{out}$ is the output of multi-head self attention layer.
	
	\begin{equation}
		\label{eq3}
		TF_{out} = MLP(NORM(MHA_{out})) + MHA_{out}
	\end{equation}
	
	Where $MLP$ is the multi layer perceptron block, and $TF_{out}$ is the output of the transformer block.

	\subsection{PlantXViT architecture}
	
\begin{figure}[!h]
		\centering
		\includegraphics[width=0.85\linewidth]{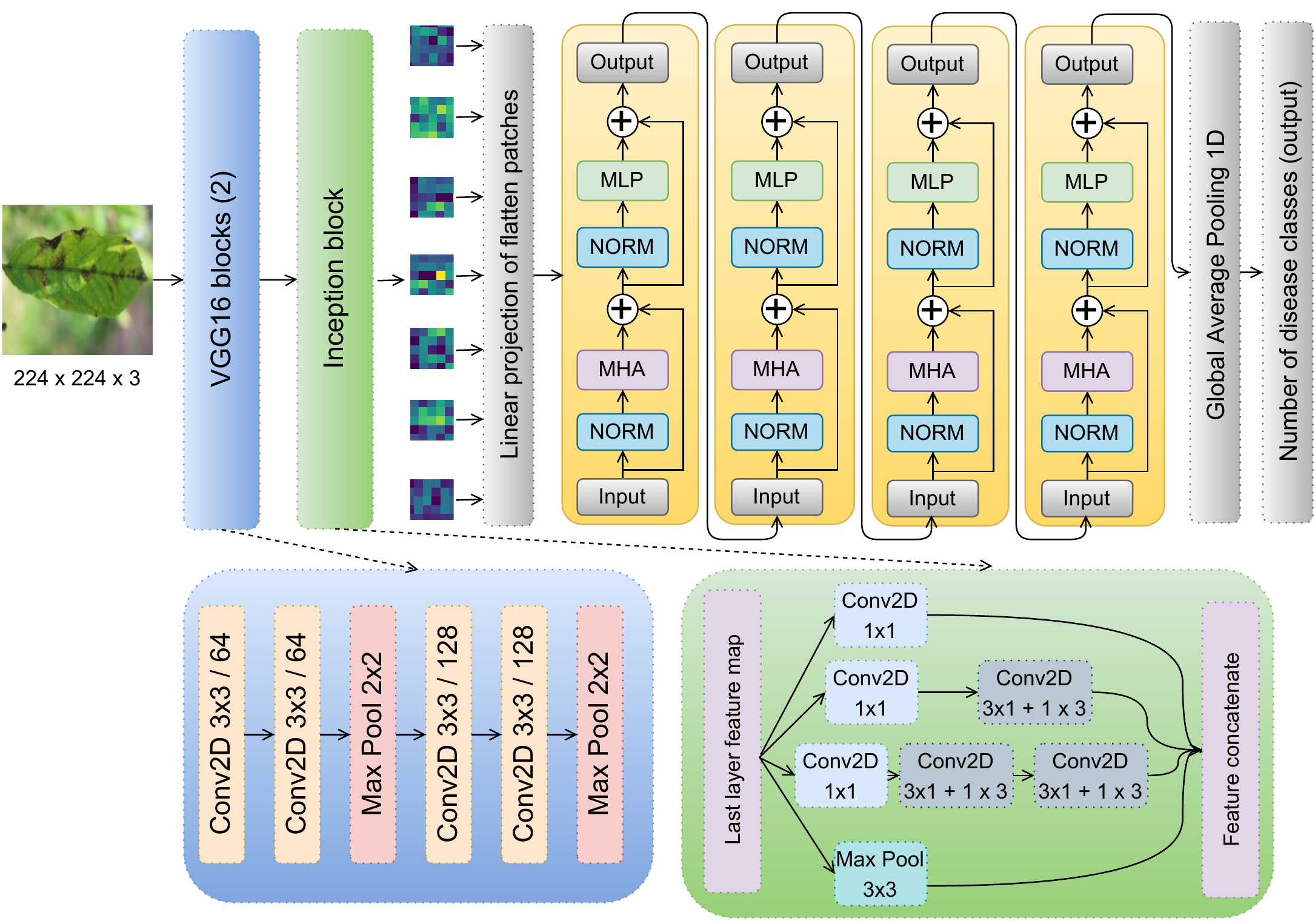}
		\caption{Block diagram of proposed PlantXViT}
		\label{fig2}
	\end{figure}

	The main objective of this work is to create a hybrid model for plant disease identification that combines the capabilities of ViT \cite{dosovitskiy2020image} and CNN for plant disease detection and identification.  Convolution ( Conv) blocks are used to efficiently extract local-level features while the transformer blocks are appended for global feature extraction. The overall pipeline of the PlantXViT is shown in Fig. \ref{fig2}. The key components of the proposed PlanxViT are  Conv blocks of VGG16 and Inception v7, and ViT components -- MHA, MLP with linear projections.

	The PlantXViT model takes an input of size $224 \times 224 \times 3$ as shown in Fig. \ref{fig2}. The model consists of two Conv blocks of VGG16 network pre-trained on the Imagenet dataset. Each block consists of two Conv layers and a max pooling layer. The output of second Conv block is $56 \times 56 \times 128$. This output is fed to a multi-level feature extraction block similar to Inception v7 Conv blocks as shown in Fig. \ref{fig2}. The multi-level feature extraction block is added for enhancing the local feature learnability of the model. The inception block generates an output of size $56 \times 56 \times 512$ after concatenating the feature maps generated by different Conv layers. 
	
	The feature map is then converted into patches, each of size $ 5 \times 5$. The flattened patches are then passed through linear projection and generate feature vector of size $121 \times 16$. These vectors are fed to a stack of four transformer blocks for features extraction. Finally, the global average pooling layer is added to convert the output of the transformer block into a 1-dimensional vector. At the end, a fully connected layer with softmax activation is added, with the number of neurons equal to the number of classes in the dataset. The layer-wise parameter count is presented in Table \ref{tab2}. For a dataset with 4 class labels, the model contains $850,500$ total trainable parameters. The model is trained, validated and tested on a variety of datasets. The experimental results on all the datasets are presented in Section \ref{result}. 
	
	\begin{table}[!h]
		\centering
		\caption{Total number of parameters of PlantXViT}
		\label{tab2}
		\scalebox{1}{
			\begin{tabular}{lrr}
				\hline
				\hline
				\textbf{Layer/Block} & \textbf{Output Shape} & \textbf{Parameters} \\
				\hline
				Input Layer &   $224 \times 224 \times 3$    &  0   \\
				Conv2D  &   $224 \times 224 \times 64$  &   1792     \\
				Conv2D  &   $224 \times 224 \times 64$  &   36,928    \\
				MaxPooling2D & $112 \times 112 \times 64$  & 0  \\
				
				Conv2D  &   $112 \times 112 \times 128$  &   73,856     \\
				Conv2D  &   $112 \times 112 \times 128$  &   147,584    \\
				MaxPooling2D & $56 \times 56 \times 128$  & 0  \\
				
				Inception v7 & $56 \times 56 \times 512$ & 361,728 \\
				PatchEncoder & $ 121\times 16$ & 206,752 \\
				
				Transformer block 1 & $ 121\times 16$  & 5440 \\
				Transformer block 2 & $ 121\times 16$  & 5440  \\
				Transformer block 2 & $ 121\times 16$  & 5440  \\
				Transformer block 4 & $ 121\times 16$  & 5440  \\
				Normalization Layer & $ 121\times 16$  & 32  \\
				Global Average Pooling 1D & $16$ & 0 \\
				Output & 4 & 68 \\
				\hline
				Total & - & \textbf{850,500} \\
				\hline
			\end{tabular}
		}
	\end{table}
	
	PlantXViT model is enriched by pre-trained VGG16 that helps in better parameter initialization; and an Inception v7 block that generates a rich pool of multi-scale features. The  transformer blocks with MHA provide an efficient mechanism for image patch processing and help extract saliency in the patches. Fusion of CNN and transformer network makes a powerful feature extractor with a combination of global and local information and enhances the explainability of the model. The model is evaluated on five public datasets and the results are presented in the next section. 
	
	\section{Results and discussion}
	\label{result}
	
	The plant disease detection model PlantXViT was developed using five publicly available datasets. Comprehensive experiments were carried out to finalize the model architecture and to evaluate its performance. Further, the model's performance was also compared with five recent CNN models \cite{karthik2020attention,chen2020attention,chen2021identifying,chen2021identification,zhao2022ric}. These include some of the recent lightweight CNN models and models that have used different attention mechanisms. In the following sections, details of the datasets, evaluation metrics, experimental setup, and performance results are presented. The results of the comparative performance of the model with other state-of-the-art CNN models are also analyzed. Further, the proposed model is evaluated in terms of its explainability using two standard methods— gradient-weighted class activation maps (Grad-CAMs) and Local Interpretable Model Agnostic Explanation (LIME). 
	
	After resizing all of the images in each dataset to $ 224 \times 224 \times 3$, the PlantXViT model was trained using training subsets from the datasets. For training the model, categorical cross-entropy loss with the Adam optimizer was used. The cross-entropy loss is defined in Eq. \ref{eq4}. The learning rate was set to 0.0001 and the batch size to 16. The validation dataset was used for the evaluation of the model's performance after each epoch. Once the model performed with the desired level of classification accuracy on the training and validation subsets, it was evaluated on the test dataset.
	
	\begin{equation}
		\label{eq4}
		Loss= -\frac{1}{n} \sum_{i=1}^{n} y_{i} \  log\hat{y}_{i}
	\end{equation}
	Here the loss for all the \emph{n} samples in a batch was calculated using $ y_{i} $ as the actual label and $ \hat{y}_{i} $ as the predicted value of the i-th sample.
	
	\subsection{Datasets}    
\begin{figure}[!h]
		\centering
		\includegraphics[width=\linewidth]{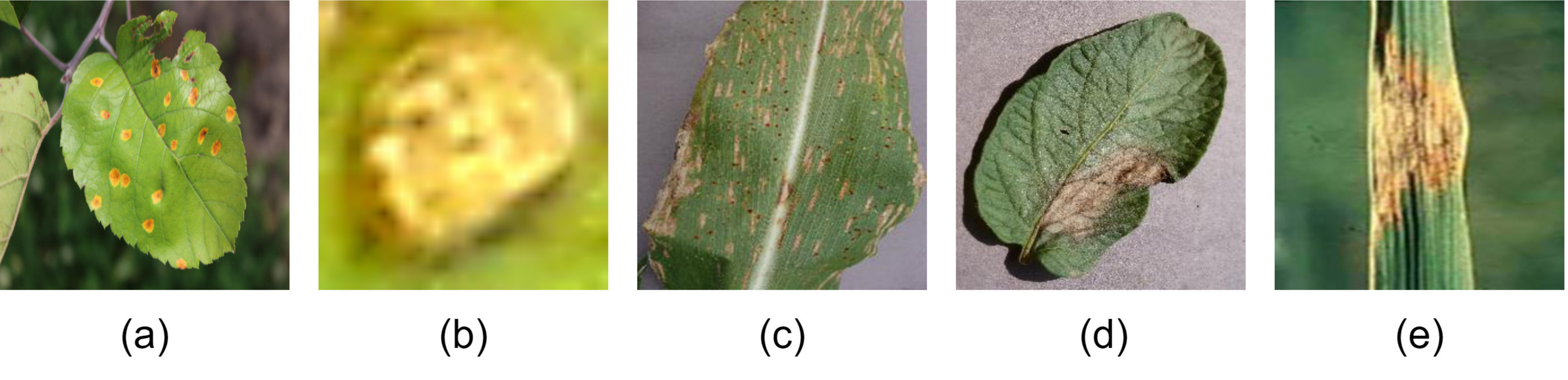}
		\caption{Sample images from all the datasets: (a) rust (Apple), (b) citrus canker (Embrapa) (c) northern leaf blight (Maize) (d) potato late blight (PlantVillage) (e) leaf scald (Rice)}
		\label{fig3}
	\end{figure}

	In the following experiments, five publicly available datasets are employed that have originated from different geographical locations and include a wide range of crops. The datasets are selected from different categories, ranging from small, balanced datasets with 400–500 images to large, imbalanced datasets with over 40K images. The goal of using these datasets from various contexts is to train the proposed model for a wide range of crops and their diseases and to evaluate its performance under various test scenarios. Out of the five publicly available datasets, PlantVillage dataset \cite{hughes2015open} contains 54,305 images in 38 classes, and Embrapa dataset \cite{barbedo2018annotated} contains 46,376 images in 93 classes. These datasets contain multiple species with a variety of diseases affecting the plant's leaves. On the other hand, Apple \cite{thapa2020plant}, Maize \cite{chen2020using} and Rice \cite{chen2020self} datasets are single-species datasets selected for the experiments. The details of each dataset with the number of classes and size are presented in Table \ref{tab3}. Sample images from each of the datasets are shown in Fig. \ref{fig3}.
	
	\begin{table}[!h]
		\centering
		\caption{Datasets used in the experiments}
		\label{tab3}
		\scalebox{1}{
			\begin{tabular}{lrr}
				\hline
				\textbf{Dataset} & \textbf{\# of classes} & \textbf{\# of images} \\
				\hline
				Apple \cite{thapa2020plant} & 4 & 1821 \\
				Embrapa \cite{barbedo2018annotated} & 93 & 46,376 \\
				Maize \cite{chen2020using} & 4 & 481 \\
				PlantVillage \cite{hughes2015open} & 38 & 54,305 \\
				Rice \cite{chen2020using} & 5 & 560 \\
				\hline
			\end{tabular}	
		}
	\end{table}
	
	\subsection{Evaluation Metrics} 
	The model evaluation is done using standard classification performance measures. These include accuracy, precision, recall, F1-score, area under the receiver operating curve, and Cohen's kappa score. 
	
	\textbf{Accuracy}: 	Accuracy is a widely used performance metric for the image classification task. It defines the relationship between the actual class value and the predicted class value. The higher the accuracy value achieved by the algorithm, the better performance is attained. Accuracy is defined as follows in Eq. \ref{eq5}.
	\begin{equation}
		\label{eq5}
		Accuracy = \frac{(TP + TN)}{(TP + TN + FP + FN)}
	\end{equation}
	
	A true positive (TP) is the count of the samples belonging to a particular class A and predicted correctly in class A. Class true negative (TN) is the count of the samples that belong to class B and are predicted as class B. False-positive (FP) is the count of the samples that belong to class B but are predicted as A, and false-negative (FN) is the count of the samples that belong to class A but are predicted as class B.
	
	\textbf{Precision}: It is the ratio of true positives to all predicted positives. It is in between the range of 0 and 1. The precision value should be as high as possible to define the preciseness of the algorithm. The formula of the precision is presented in Eq. \ref{eq6}.
	
	\begin{equation}
		\label{eq6}
		Precision = \frac{TP}{(TP + FP)}
	\end{equation}
	
	\textbf{Recall}: It defined the ratio between the true positive labels to all the actual positive labels. The value of recall lies in between 0 to 1. It is calculated as shown in Eq. \ref{eq7}.
	\begin{equation}
		\label{eq7}
		Recall = \frac{TP}{(TP + FN)}
	\end{equation}
	
	\textbf{F1-Score}: 	It relates Precision and Recall by calculating the harmonic mean between them. The formula for F1-score is depicted in Eq. \ref{eq8}.
	\begin{equation}
		\label{eq8}
		F1-Score = 2 * \frac{(Precision * Recall)}{(Precision + Recall)}
	\end{equation}
	
	\textbf{Area Under the Curve (AUC)}: AUC is the area covered by the receiver characteristic operator (ROC) curve. The ROC is calculated using the plot of true positive rate (TPR) (refer Eq. \ref{eq9}) and false positive rate (FPR) (refer Eq. \ref{eq10}). 
	\begin{equation}
		\label{eq9}
		TPR = \frac{TP}{(TP + FN)}
	\end{equation}
	\begin{equation}
		\label{eq10}
		FPR = \frac{FP}{(FP + TN)}
	\end{equation}
	
	\textbf{Cohen's Kappa Score}: Cohen's kappa coefficient or score is a probability based measure where the outcome is the level of agreement parties for classification problem. The formula of kappa score calculation is shown below in Eq. \ref{eq11}.
	
	\begin{equation}
		\label{eq11}
		Kappa = \frac{p_o-p_e}{(1-p_e)}
	\end{equation}
	
	Where $p_o$ is the relative agreement probability and $p_e$ is the hypothetical agreement probability  between the parties. 
	
	In addition, confusion matrix, receiver-operating characteristic (ROC) curve analysis, t-distributed stochastic neighbor embedding (t-SNE) plots are used for evaluating the performance of the model. The confusion matrix and ROC curve indicate the credibility of the model. The higher the ROC curve in the upper left corner, the better is the model's performance. The t-SNE plots demonstrated how good the model is in terms of distinctive feature extraction for different categories in a dataset \cite{van2008visualizing}.	Further, the model's explainability is evaluated using two standard methods, namely Grad-CAMs  and LIME.
	
	\subsection{Experimental setup}
	
	All the experiments were performed on an Nvidia DGX A100 160GB station with four GPU A100 cards, each with 40 GB of memory. It has the Ubuntu 18.04 LTS operating system on the machine, with an AMD 7742 processor at 2.25–3.4 GHz and 512 GB of RAM. The proposed model and other selected models for comparison were implemented using the Keras framework with NVIDIA CUDA v11.5 and the cuDNN v8.3 library.
	
	\subsection{Results and analysis}  
	
	A series of comparative experiments were carried out using CNNs and attention models on the test sets in order to show the effectiveness of the PlantXViT model in the plant disease detection task. One of the experiments is evaluated to select the patch size to feed it into the ViT part of the PlantXViT model. The size of each patch is a hyper-parameter which needs to be selected carefully to generate the best results. In the work, we experimented different patch sizes and came up with the best among them. The patch sizes of 1, 3, 5, 7, and 9 were selected in the experiment. The results for all the patch sizes are presented in Table \ref{tab4}. As per the results, patch size of 5 generated the best results in terms of accuracy, precision, recall and f1-score for all the datasets. However, AUC for patch size 1 and patch size 7 are better in Apple and Rice datasets, respectively. 
	
	\begin{table}[!h]
		\centering
		\caption{PlantXViT's performance with different patch size}
		\label{tab4}
		\scalebox{1}{
			\begin{tabular}{lrrrrrrr}
				\hline
				\hline
				\textbf{Patch size} & \textbf{Loss} & \textbf{Accuracy} & \textbf{Precision} & \textbf{Recall} & \textbf{F1 score} & \textbf{AUC} & \textbf{kappa score} \\
				\hline
				\hline
				\multicolumn{8}{c}{\textbf{Apple}} \\
				\hline
				1 & 0.31 & 92.47 & 92.47 & 92.47 & 92.47 & \textbf{97.61} & 0.89 \\
				3 & 0.58 & 81.72 & 81.52 & 80.65 & 81.08 & 95.21 & 0.73 \\
				5 & \textbf{0.3} & \textbf{93.55} & \textbf{93.55} & \textbf{93.55} & \textbf{93.55} & 97.01 & \textbf{0.91} \\
				7 & 0.44 & 89.25 & 89.73 & 89.25 & 89.49 & 96.91 & 0.84 \\
				9 & 0.43 & 88.17 & 88.11 & 87.63 & 87.87 & 96.83 & 0.83 \\
				\hline
				\multicolumn{8}{c}{\textbf{Embrapa}} \\
				\hline
				1 & 0.53 & 87.25 & 89.59 & 85.85 & 87.68 & 98.52 & 0.87 \\
				3 & 0.54 & 86.61 & 88.64 & 85.11 & 86.84 & 98.44 & 0.86 \\
				5 & \textbf{0.46}  & \textbf{89.24} & \textbf{91.17} & \textbf{88.27} & \textbf{89.7} & \textbf{98.73} & \textbf{0.89} \\
				7 & 0.52 & 87.63 & 89.49 & 86.68 & 88.06 & 98.25 & 0.87 \\
				9 & 0.68 & 84.73 & 86.81 & 83.5 & 85.12 & 97.72 & 0.84 \\
				\hline
				\multicolumn{8}{c}{\textbf{Maize}} \\
				\hline
				1 & 0.42   & 86.42 & 88.61 & 86.42 & 87.5 & 96.74 & 0.82 \\
				3 & 0.44 & 90.12 & 91.03 & 87.65 & 89.3 & 95.48 & 0.87 \\
				5 & \textbf{0.35} & \textbf{92.59} & \textbf{92.5} & \textbf{91.36} & \textbf{91.93} & \textbf{96.8} & \textbf{0.9} \\
				7 & 0.44 & 88.89 & 92.31 & 88.89 & 90.57 & 95.43 & 0.85 \\
				9 & 0.42 & 87.65 & 88.75 & 87.65 & 88.2 & 96.76 & 0.84 \\
				\hline
				\multicolumn{8}{c}{\textbf{PlantVillage}} \\
				\hline
				1 & 0.11 & 97.06 & 97.38 & 96.86 & 97.12 & 99.76 & 0.97 \\
				3 & 0.05 & 98.66 & 98.71 & 98.62 & 98.66 & 99.87 & \textbf{0.99} \\
				5 & \textbf{0.04} & \textbf{98.86} & \textbf{98.9} & \textbf{98.81} & \textbf{98.85} & \textbf{99.92} & \textbf{0.99} \\
				7 & 0.09 & 97.3 & 97.58 & 97.08 & 97.33 & 99.82 & 0.97 \\
				9 & 0.08 & 97.85 & 98.01 & 97.82 & 97.91 & 99.78 & 0.98 \\
				\hline
				\multicolumn{8}{c}{\textbf{Rice}} \\
				\hline
				1 & 0.25 & 93.33 & 94.92 & \textbf{93.33} & 94.12 & 99.41 & 0.91 \\
				3 & 0.27   & 91.67 & 94.74 & 90    & 92.31 & 99.38 & 0.89 \\
				5 & 0.24 & \textbf{95} & \textbf{98.25} &\textbf{93.33} & \textbf{95.73} & 99.39 & \textbf{0.94} \\
				7 & \textbf{0.21} & 93.33 & 94.83 & 91.67 & 93.22 & \textbf{99.73} & 0.91 \\
				9 & \textbf{0.21} & 95    & 94.92 & \textbf{93.33} & 94.12 & 99.51 & \textbf{0.94} \\
				\hline
				\hline
			\end{tabular}
		}
	\end{table}
	
	Performances of several optimizers were also evaluted for PlantXViT model to select the best among all for improved training. In the experiments, SGD, RMSProp, Adamax, Adam, and Nadam optimizers were compared to ensure the effectiveness of the model. According to the results in Table \ref{tab5}, the PlantXViT model with the SGD optimizer showed a substantial decrease in the model results, while Adam was able to maintain the accuracy across all the datasets. Further, the Nadam optimizer performed well on PlantVillage dataset. However, the Adam optimizer's performance was found to be consistent across all the five datasets, irrespective of their sizes.
	
	\begin{table}[!h]
		\centering
		\caption{Comparison of the various optimizer}
		\label{tab5}
		\scalebox{1}{
			\begin{tabular}{lrrrrrrr}
				\hline
				\hline
				\textbf{Optimizer} & \textbf{Loss} & \textbf{Accuracy} & \textbf{Precision} & \textbf{Recall} & \textbf{F1 score} & \textbf{AUC} & \textbf{kappa score} \\
				\hline
				\hline
				\multicolumn{8}{c}{\textbf{Apple}}\\
				\hline
				SGD  & 0.95 & 60.22 & 68.25 & 46.24 & 55.13 & 84.36 & 0.42 \\
				RMSProp & 0.35 & \textbf{93.55} & \textbf{93.55} & \textbf{93.55} & \textbf{93.55}    & 97.34 & 0.91 \\
				Adamax & 0.81 & 68.28 & 73.86 & 60.75 & 66.67 & 88.66 & 0.54 \\
				Adam & 0.3 & \textbf{93.55 } & \textbf{93.55} & \textbf{93.55} & \textbf{93.55} & 97.01 & 0.1 \\
				Nadam & \textbf{0.28} & 93.01 & 93.51 & 93.01 & 93.26 & \textbf{97.89} & \textbf{0.9} \\
				
				\hline
				\multicolumn{8}{c}{\textbf{Embrapa}}\\
				\hline
				SGD       & 0.72 & 81.31    & 91.72     & 72.42  & 80.94         & 98.99 & 0.81 \\
				RMSProp   & 0.71 & 83.33    & 83.33     & 80.59  & 81.94         & 97.83 & 0.83  \\
				Adamax    & 0.56 & 86       & 92.28     & 81.16  & 86.36         & 98.99 & 0.85 \\
				Adam      & \textbf{0.46} & \textbf{89.24} & \textbf{91.17} & \textbf{88.27} & \textbf{89.7} & \textbf{98.73} & \textbf{0.89} \\
				Nadam     & 0.53 & 87.04    & 88.72     & 85.77  & 87.22         & 98.41 & 0.87 \\
				
				\hline
				\multicolumn{8}{c}{\textbf{Maize}}\\
				\hline
				SGD       & 0.75 & 70.37    & 73.68     & 69.14  &  71.34        & 90.32 & 0.61 \\
				RMSProp   & 0.44  & 87.65    & 88.31     & 83.95  & 86.07         & 96.04 & 0.84 \\
				Adamax    & 0.44 & 90.12    & \textbf{92.11}     & 86.42  & 89.17         & 96.1 & 0.87 \\
				Adam      & \textbf{0.35} & \textbf{92.59}    & 92.5     & \textbf{91.36}  &  \textbf{91.93}        & \textbf{96.8} & \textbf{0.9} \\
				Nadam     & 0.47 & 87.65    & 90.79     & 85.19  &  87.9        & 94.99 & .83 \\

				\hline
				\multicolumn{8}{c}{\textbf{PlantVillage}}\\
				\hline
				SGD       & 0.09 & 97.5     & 98.07     & 96.79  & 97.43          & 99.95 & 0.97 \\
				RMSProp   & 0.08 & 97.72    & 97.94     & 97.52  & 97.73         & 99.80 & 0.98 \\
				Adamax    & 0.06 & 98.49    & 98.58     & 98.33  &  98.45        & 99.91 & 0.98 \\
				Adam      & 0.04 & 98.86    & 98.9     & 98.81  & 98.85         & 99.92 & 0.99 \\
				Nadam     & \textbf{0.03} & \textbf{99.06} & \textbf{99.14} & \textbf{98.99} & \textbf{99.06} & \textbf{99.92} & \textbf{0.99} \\
				
				\hline
				\multicolumn{8}{c}{\textbf{Rice}}\\
				\hline
				SGD       & 0.98  & 70       & 83.78     & 51.67  &  63.92        & 88.8  & 0.63 \\
				RMSProp   & 0.36 & 91.67    & 94.55     & 86.67  &   90.44       & 98.38 & 0.89 \\
				Adamax    & 0.55 & 85       & 94        & 78.33  &  85.45        & 96.6  & 0.81 \\
				Adam & \textbf{0.24} & \textbf{95} & \textbf{98.25} & \textbf{93.33} & \textbf{95.73} & \textbf{99.39} & \textbf{0.94} \\
				Nadam     & 0.66 & 76.67    & 80.36     & 75     & 77.59         & 94.31 & 0.71 \\
				\hline
				\hline
			\end{tabular}
		}
	\end{table}
	
\begin{figure}[!h]
		\centering
		\includegraphics[height=0.93\textheight]{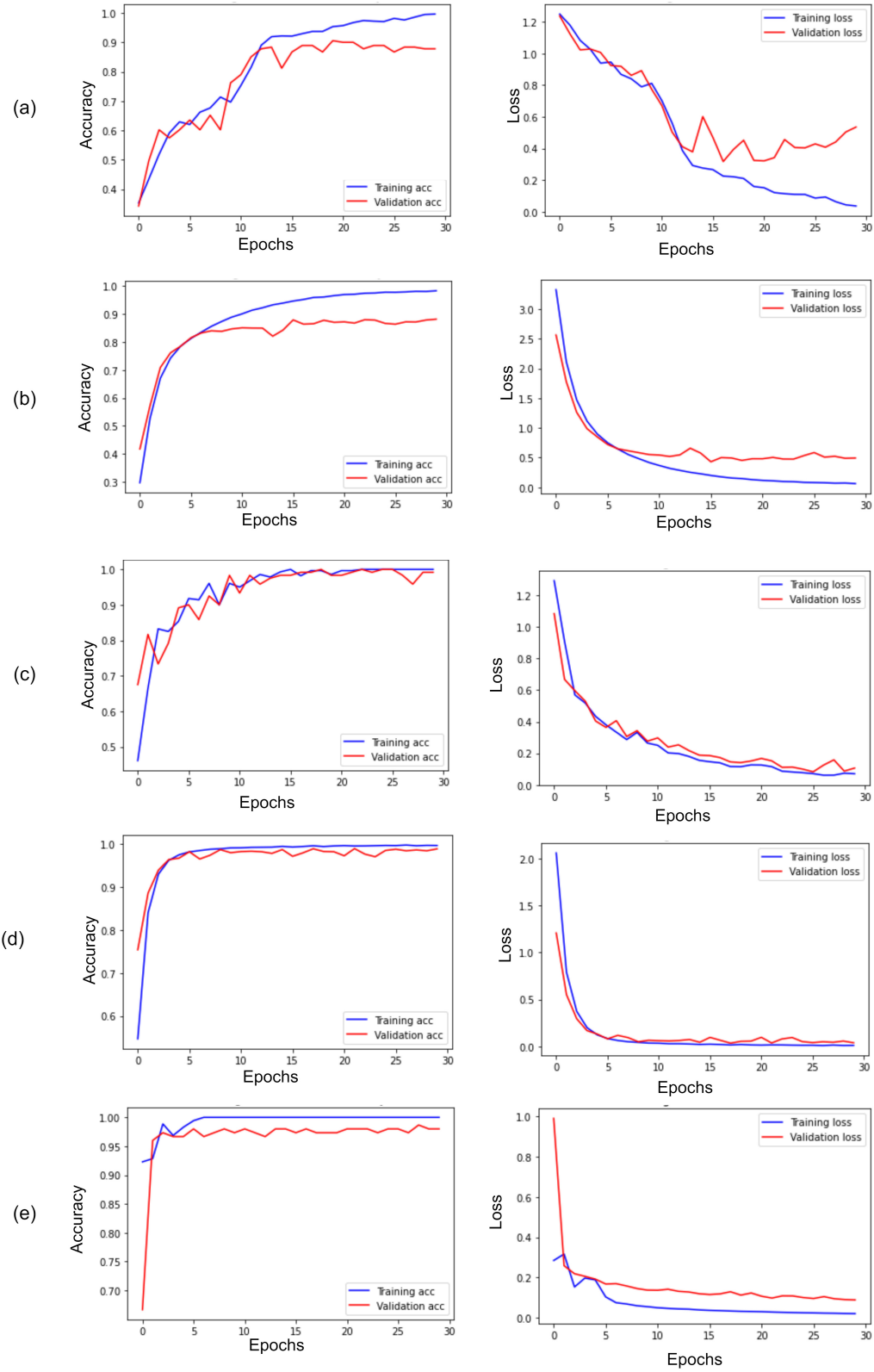}
		\caption{Accuracy and loss graph for (a) Apple, (b) Embrapa, (c) Maize, (d) PlantVillage, and (e) Rice datasets.}
		\label{fig4}
	\end{figure}

	The training and validation epoch-wise graph of accuracy and loss is presented in Fig. \ref{fig4} for all the five datasets. The graphs converge perfectly for Maize, PlantVillage, and Rice datasets, whereas there is more difference in the training and validation data results for Apple and Embrapa datasets. For Embrapa dataset, the reason for this difference could be high imbalance and very few samples in case of some classes. In Apple dataset, samples with multiple diseases get misclassified due to textural similarity between different type of diseases.
	
	\begin{figure}[!h]
		\centering
		\includegraphics[width=0.85\linewidth]{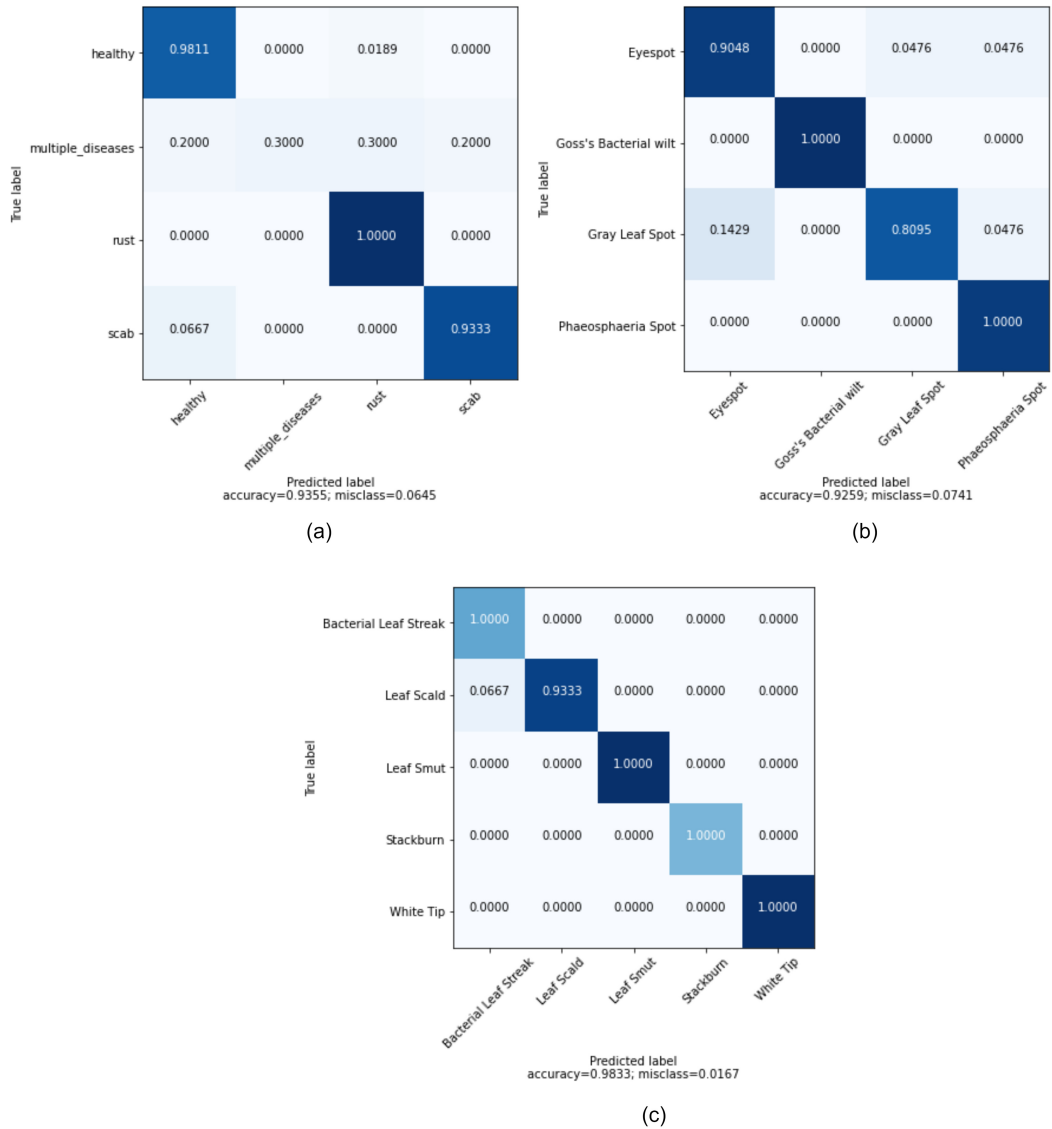}
		\caption{Normalized confusion matrix for (a) Apple, (b) maize, and (c) Rice datasets. }
		\label{fig5}
	\end{figure}
	
\begin{figure}[!h]
		\centering
		\includegraphics[height=0.9\textheight]{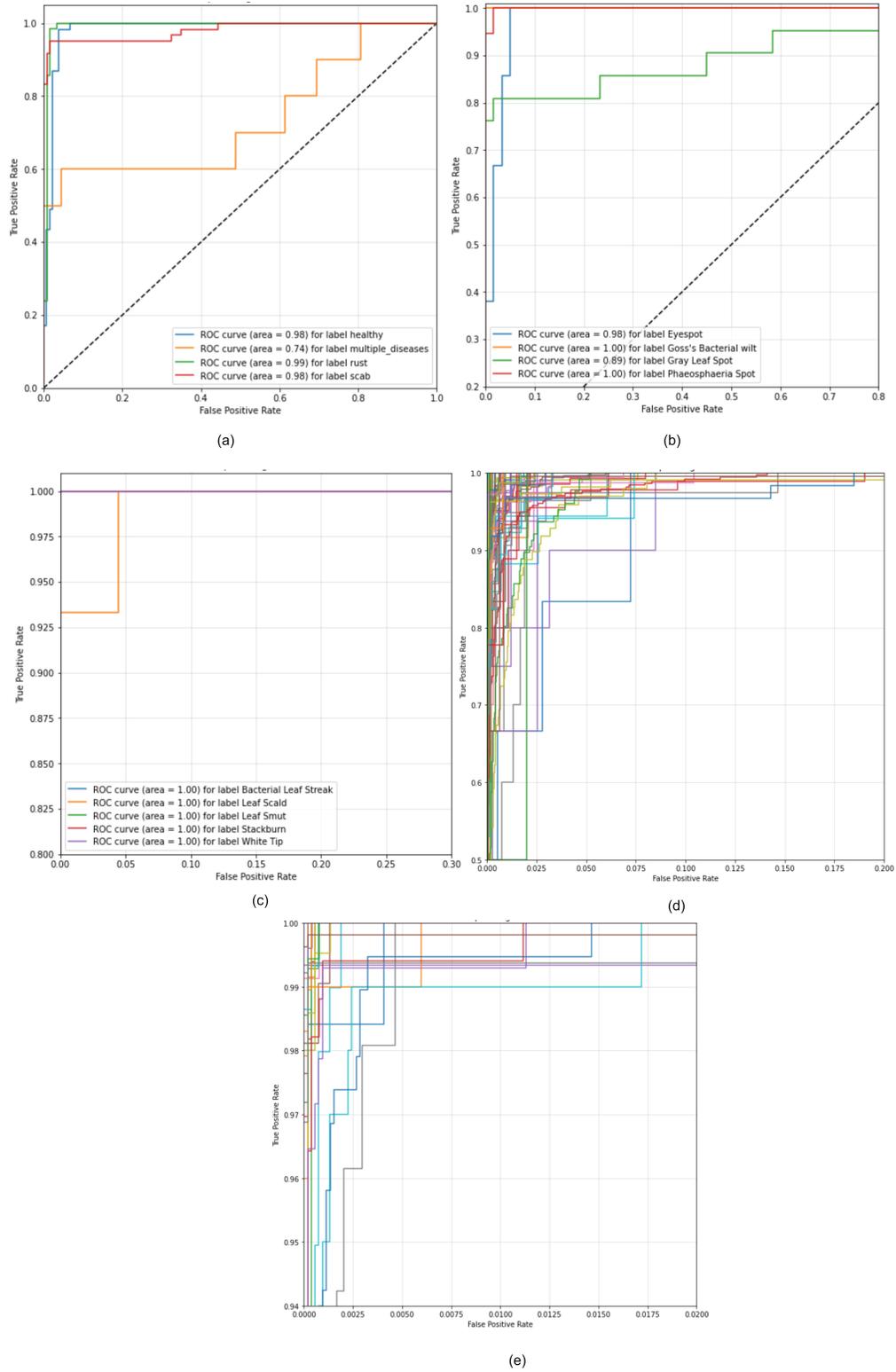}
		\caption{ROC for (a) Apple, (b) Maize, (c) Rice, (d) Embrapa, and (e) PlantVillage datasets. }
		\label{fig6}
	\end{figure}

	The confusion matrices of the test set of Apple, Maize, and Rice datasets using PlantXViT are shown in Fig. \ref{fig5}. As the other two datasets, PlantVillage and Embrapa, have a large number of classes, it is not possible to plot the confusion matrix for the same. It can be observed that PlantXViT has a higher classification accuracy in all the three datasets. In the confusion matrix Fig. \ref{fig5} (a), we can see that the classifications of healthy, rust, and scab are very accurate for Apple dataset. However, multiple diseases are highly misclassified, with only 30\% of them correctly classified. Similarly, in Maize dataset (Fig. \ref{fig5} (b)), Goss’s bacterial wilt and phaeosphaeria spot are classified correctly. However, 10\% of the test images in the eyespot class are misclassified as gray leaf spot and phaeosphaeria spot. On the other hand, 20\% of the gray leaf images are misclassified as eye spots and phaeosphaeria spots. In Rice (see Fig. \ref{fig5} (c)) dataset, the model correctly classifies images of bacterial leaf streak, leaf smut, stackburn, and white tip. Only 6.67\% of the leaf scald images are misclassified as bacterial leaf streaks. It can be observed that leaves having more than one disease are highly misclassified in Apple dataset. On the other hand, diseases having similar structural elements, i.e., spots, are misclassified. The ROC curve for all five datasets is shown in Fig.\ref{fig6}. The aforementioned figure demonstrates that the PlantXViT has achieved a good performance in all the five datasets. 
	
	PlantXViT is compared with five recently introduced CNN-based plant disease classification models \cite{karthik2020attention,chen2020attention,chen2021identifying,chen2021identification,zhao2022ric}. The models chosen for comparison in this work are mainly recent models with attention mechanisms and lightweight structures. Each model represents a unique technique for disease detection, which makes the comparative results instructive. The parameters of the models are set to the same as the PlantXViT parameters for training. Table \ref{tab6} shows the quantitative results of the different CNN models on all the five datasets. The PlantXViT model has an accuracy of 93.55\%, 89.24\%, 92.59\%, 98.86\%, and 98.33\% on Apple, Embrapa, Maize, PlantVillage, and Rice datasets, respectively. Table \ref{tab6} illustrates that the CNN models \cite{karthik2020attention,chen2020attention,chen2021identifying,chen2021identification,zhao2022ric} designed on the basis of the attention mechanism are less effective than PlantXViT. It is able to achieve the best performance in all metrics for all five datasets. 
	
	\begin{table}[!h]
		\centering
		\caption{Comparison of the work with other state-of-the-art methods}
		\label{tab6}
		\scalebox{1}{
			\begin{tabular}{lrrrrrrr}
				\hline
				\hline
				\textbf{Approach} & \textbf{Loss} & \textbf{Accuracy} & \textbf{Precision} & \textbf{Recall} & \textbf{F1 score} & \textbf{AUC} & \textbf{kappa score} \\
				\hline
				\hline
				
				\multicolumn{8}{c}{\textbf{Apple}}\\
				\hline
				\citet{karthik2020attention} & 1.56 & 62.37 & 62.84 & 61.83 & 62.35 & 83.53 & 0.45 \\
				
				\citet{chen2020attention} & 1.96 & 55.91 & 56.35 & 54.84 & 55.58 & 78.32 & 0.36 \\
				
				\citet{chen2021identifying} & 0.79 & 83.33 & 83.7 & 82.8 & 83.25 & 94.1 & 0.76 \\
				
				\citet{chen2021identification} & 0.63 & 83.33 & 85.47 & 82.26 & 83.83 & 95.59 & 0.76 \\
				
				\citet{zhao2022ric} & 0.58 & 88.71 & 89.62 & 88.17 & 88.89 & 95.81 & 0.84 \\
				
				\textbf{PlantXViT} & \textbf{0.3} & \textbf{93.55} & \textbf{93.55} & \textbf{93.55} & \textbf{93.55} & \textbf{97.01} & \textbf{0.91} \\
				\hline
				\multicolumn{8}{c}{\textbf{Embrapa}}\\
				\hline
				\citet{karthik2020attention} & 0.77 & 80.29 & 83.07 & 78.38  & 80.6 & 97.8 & 0.82 \\
				
				\citet{chen2020attention} & 0.6 & 80.95 & 85.08 & 76.02 & 80.3 & 98.02 & 0.78 \\
				
				\citet{chen2021identifying} & 1.11 & 73.63 & 80.12 & 67.86 & 73.48 & 98.02 & 0.73 \\
				
				\citet{chen2021identification} & 1.12 & 74.88  & 82.93 & 66.06 & 73.18  & 98.2 & 0.75 \\
				
				\citet{zhao2022ric} & 0.36 & 88.48 & 89.13 & 87.44 & 88.28 & 98.23 & 0.88 \\
				
				\textbf{PlantXViT} & \textbf{0.46} & \textbf{89.24} & \textbf{91.17} & \textbf{88.27} & \textbf{89.7} & \textbf{98.73} & \textbf{0.89} \\
				\hline
				\multicolumn{8}{c}{\textbf{Maize}}\\
				\hline
				\citet{karthik2020attention} & 1.17 & 54.32 & 59.72 & 53.09 & 56.21 & 83.19 & 0.39 \\
				
				\citet{chen2020attention} & 1.26 & 49.38 & 50.85 & 37.04 & 42.86 & 76.16 & 0.33 \\
				
				\citet{chen2021identifying} & 0.6 & 87.65 & 87.65 & 87.65 & 87.65 & 96.03 & 0.84 \\
				
				\citet{chen2021identification} & 0.5 & 88.89 & 91.03 & 87.65 & 89.31 & 96.78 & 0.85 \\
				
				\citet{zhao2022ric} & 1.4 & 77.78 & 78.75 & 77.78 & 78.26 & 94.55 & 0.7 \\
				
				\textbf{PlantXViT} & \textbf{0.34} & \textbf{92.59} & \textbf{93.67} & \textbf{91.36} & \textbf{92.5} & \textbf{97.21} & \textbf{0.9} \\
				\hline
				\multicolumn{8}{c}{\textbf{PlantVillage}}\\
				\hline
				\citet{karthik2020attention} & 0.16 & 95.83 & 96.2 & 95.6 & 95.89 & 99.7 & 0.96  \\
				
				\citet{chen2020attention} & 0.4 & 87.94 & 89.59 & 86.71 & 88.07 & 99.14 & 0.85 \\
				
				\citet{chen2021identifying} & 1.07 & 74.63 & 80.92 & 69.64 & 75.03 & 98.18 & 0.74 \\
				
				\citet{chen2021identification} & 0.17 & 96.68 & 97.49 & 95.83 & 96.64 & 99.26 & 0.97 \\
				
				\citet{zhao2022ric} & 0.12 & 97.28 & 97.49 & 97.06 & 97.27 & 99.78 & 97.15 \\
				
				\textbf{PlantXViT} & \textbf{0.04} & \textbf{98.86} & \textbf{98.90} & \textbf{98.81} & \textbf{98.85} & \textbf{99.92} & \textbf{0.99} \\
				\hline
				\multicolumn{8}{c}{\textbf{Rice}}\\
				\hline
				\citet{karthik2020attention} & 0.99 & 71.67 & 79.55 & 58.33 & 69.31 & 88.26 & 0.64 \\
				
				\citet{chen2020attention} & 0.75 & 76.67 & 83.02 & 73.33 & 77.87 & 92.93 & 0.71 \\
				
				\citet{chen2021identifying} & 0.3 & 96.67 & 96.67 & 96.67 & 96.67 & 98.97 & 0.95 \\
				
				\citet{chen2021identification} & 0.25 & 95 & 96.61 & 96.61 & 96.61 & 99.72 & 0.94  \\
				
				\citet{zhao2022ric} & 0.41 & 91.67 & 91.67 & 91.67 & 91.67 & 98.55 & 89.27 \\
				
				\textbf{PlantXViT} & \textbf{0.08} & \textbf{98.33} & \textbf{98.33} & \textbf{98.33} & \textbf{98.33} & \textbf{99.94} & \textbf{0.98} \\
				\hline
				\hline
			\end{tabular}
		}
	\end{table}
	
	\begin{figure}[!h]
		\centering
		\includegraphics[height=0.9\textheight]{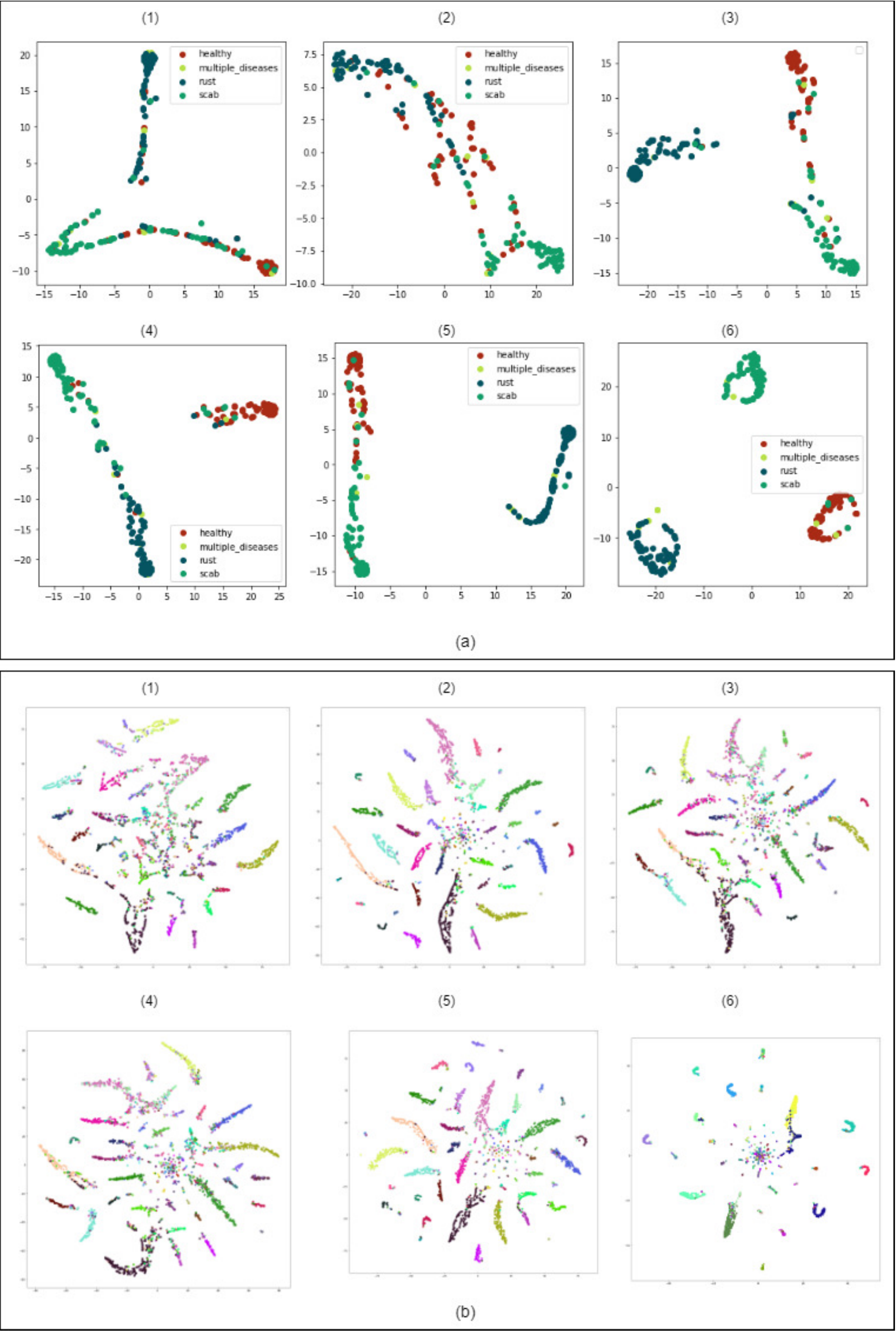}
		\caption{t-SNE plots for (a) Apple and (b) Embrapa datasets. (1) \citet{karthik2020attention} (2) \citet{chen2020attention} (3) \citet{chen2021identifying} (4) \citet{chen2021identification} (5) \citet{zhao2022ric} (6) PlantXViT}
		\label{fig7}
	\end{figure}
	\begin{figure}[!h]
		\centering
		\includegraphics[height=0.9\textheight]{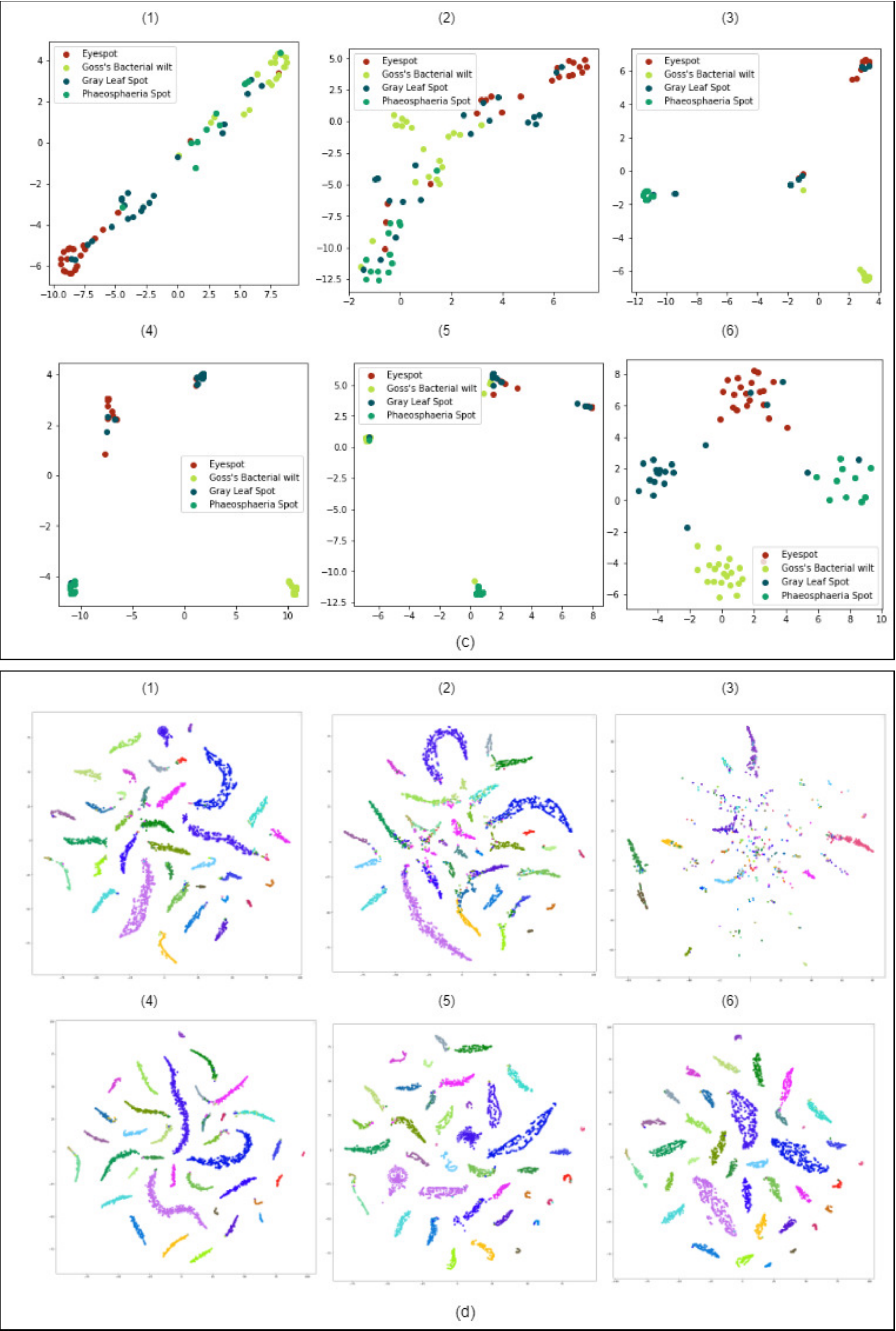}
		\caption{t-SNE plots for (c) Maize and (d) PlantVillage datasets. (1) \citet{karthik2020attention} (2) \citet{chen2020attention} (3) \citet{chen2021identifying} (4) \citet{chen2021identification} (5) \citet{zhao2022ric} (6) PlantXViT }
		\label{fig8}
	\end{figure}
	\begin{figure}[!h]
		\centering
		\includegraphics[width=0.85\linewidth]{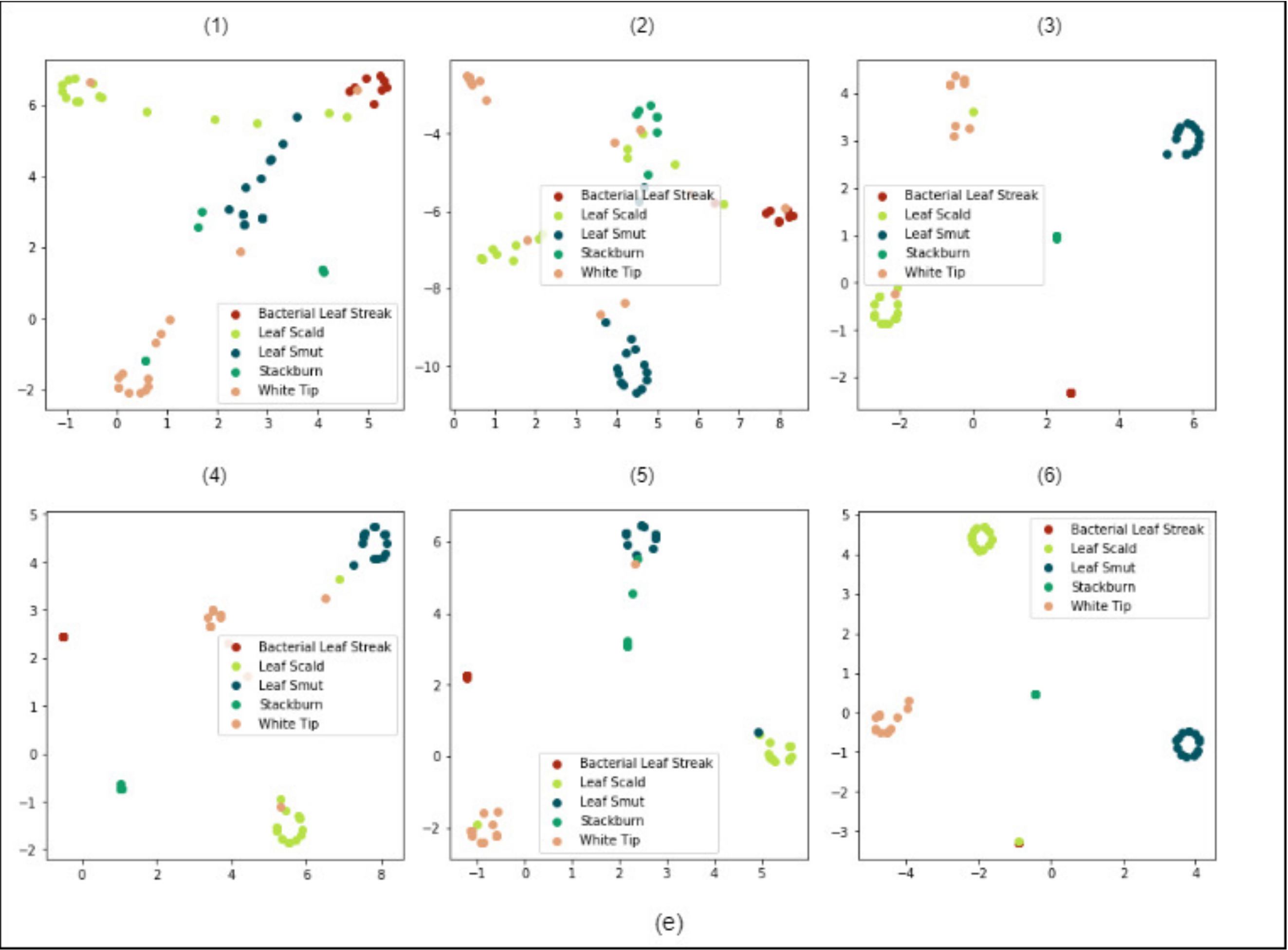}
		\caption{t-SNE plots for (e) Rice dataset. (1) \citet{karthik2020attention} (2) \citet{chen2020attention} (3) \citet{chen2021identifying} (4) \citet{chen2021identification} (5) \citet{zhao2022ric} (6) PlantXViT }
		\label{fig9}
	\end{figure}
	
	\begin{figure}[!h]
		\centering
		\includegraphics[width=0.9\linewidth]{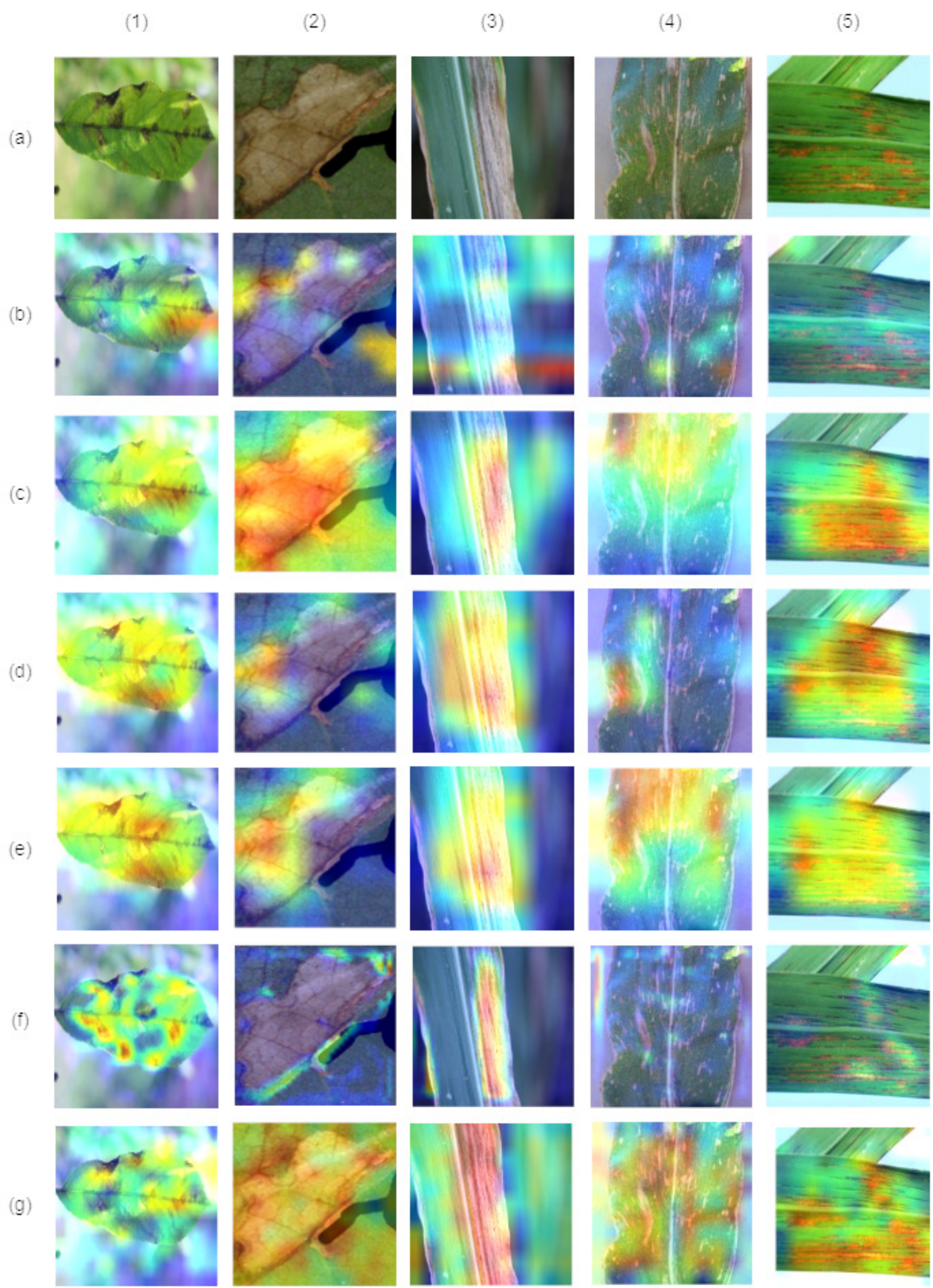}
		\caption{Grad-CAMs for (1) Apple, (2) Embrapa, (3) Maize, (4) PlantVillage, and (5) Rice datasets. (a) input image (b) \citet{karthik2020attention} (c) \citet{chen2020attention} (d) \citet{chen2021identifying} (e) \citet{chen2021identification} (f) \citet{zhao2022ric} (g) PlantXViT}
		\label{fig10}
	\end{figure}
	
	\begin{figure}[!h]
		\centering
		\includegraphics[width=0.9\linewidth]{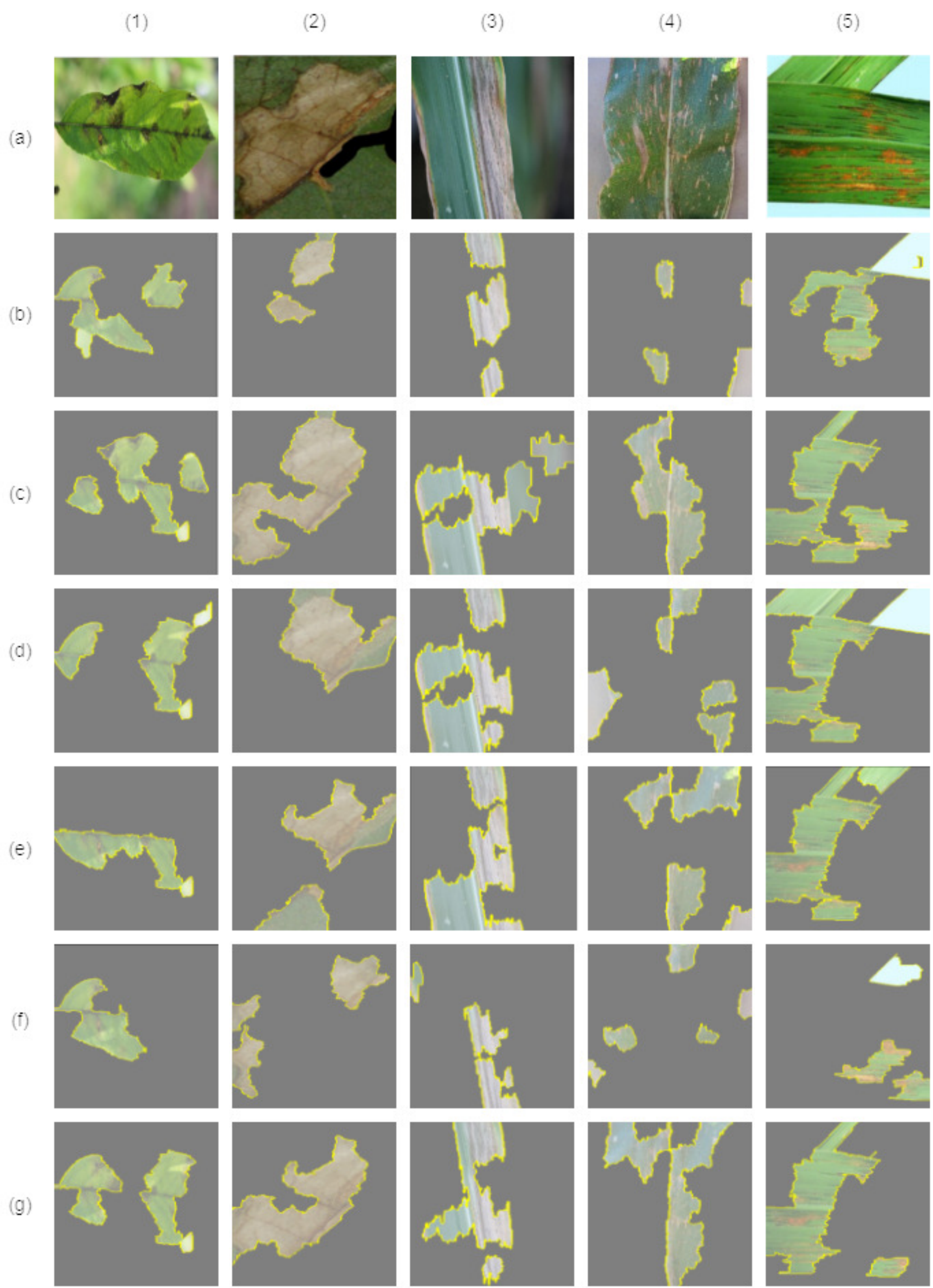}
		\caption{LIME for (1) Apple, (2) Embrapa, (3) Maize, (4) PlantVillage, and (5) Rice datasets. (a) input image (b) \citet{karthik2020attention} (c) \citet{chen2020attention} (d) \citet{chen2021identifying} (e) \citet{chen2021identification} (f) \citet{zhao2022ric} (g) PlantXViT}
		\label{fig11}
	\end{figure}
	
	In order to explain the effectiveness of PlantXViT in extracting right features for plant disease detection, the t-SNE method is used that shows the feature similarity and dissimilarity for samples of the same and different classes in a dataset \cite{van2008visualizing}.  The learned features of different datasets are  projected onto the 2D-plane using t-SNE algorithm, an unsupervised nonlinear dimensionality reduction technique \cite{van2008visualizing}. The visualisation results for all five datasets using 2-D vector scatter plots obtained with the t-SNE method are shown in Figures \ref{fig7}, \ref{fig8}, and \ref{fig9}. The results are also compared with the t-SNE plots of other state-of-the-art techniques. Output of the global average pooling layer is used to visualise feature maps for the proposed model. In Apple dataset  (refer Fig. \ref{fig7} (a)), red color denotes the healthy class, lemon green denotes multiple diseases in a single image, blue denotes the rust disease, and green denotes the scab disease. The t-SNE features for all state-of-the-art methods have been compared to analyze distinctive feature extraction capabilities of different methods. Fig. \ref{fig7} (a)(6) shows that the features are easily distinguishable and are the best among all other methods for three disease classes, while categorization is difficult for multiple disease classes. Similarly, from Fig. \ref{fig7} (b)(6) it can be observed that the proposed method can generate better clusters than all other approaches for Embrapa dataset. In Fig. \ref{fig8} (c)(6), it is visible that all the four clusters are more separable than those for the other five methods for Maize dataset. Similarly, Fig. \ref{fig8} (d)(6) demonstrates that the proposed method produces feature clusters comparable to all other methods. Fig. \ref{fig9} (e) shows the clustering capabilities of the proposed method on Rice dataset. Thus, it can be concluded that the proposed model PlantXViT is quite efficient in understanding the salient features of different classes in a dataset. 
	
	Grad-CAM \cite{selvaraju2017grad} results for the PlantXViT model and other competing models are shown in Fig. \ref{fig10} using a sample of each of the five datasets. It is worth noting that the model developed by \citet{karthik2020attention} not so incapable in identifying the correct disease portion for all the five datasets, while other models identify the portion from the entire leaf with greater precision in Apple, Maize, and Rice datasets (\cite{chen2020attention,chen2021identification,chen2021identifying}. Further, the model by \citet{zhao2022ric} is able to generate the structure of the disease portion in Apple and Maize datasets. However, it is unable to produce good results for Embrapa, PlantVillage, and Rice datasets. The PlantXViT is able to focus on the disease portion with better understanding of the disease texture and shape. These results have been verified for a large number of samples from the test datasets. For brevity, only one sample is chosen from each dataset to show representative results.

	LIME is another interpretability method that uses model-agnostic features for analyzing the interpretability of results for a classifier \cite{ribeiro2016should}. It is based on the local linear approximation of the model. The LIME results for the PlantXViT model and other comparative models are presented in Fig. \ref{fig11}. The first row (a) shows the input image from each dataset. It may be noted that the model by \citet{karthik2020attention} is not able to identify the disease area correctly for the sample from PlantVillage dataset. Similarly, the model by \citet{chen2021identifying} is not able to identify the disease correctly. The model by \citet{chen2020attention} generates comparatively better results. Further, the model by \citet{chen2021identification} generates good results for apple and Rice datasets, but not for other samples. However, the LIME results for \citet{zhao2022ric} indicate that the results are not much explainable for all the five datasets. It is indeed remarkable that the proposed PlantXViT model is able to capture the disease portion with better precision in the case of all the five datasets. These samples are just representative of how the methods are interpreting different types of diseases. These results have been verified for a large number of samples from the test datasets. 
	
	The PlantXViT model is designed with a lower number of trainable parameters to make the model compatible with mobile/handheld devices like drones and smartphones. A model's suitability for IoT devices does only depends on its low memory footprint, but also heavily depends on the computational needs, like the floating point operations (FLOPs). Table \ref{tab7} shows the total number of parameters, trainable parameters, and GegaFLOPs ( GFLPOs). It may be noted that the proposed model has reasonably lower number of trainable and total parameters as compared to other models except for the case of the model developed by \citet{karthik2020attention}. But the downside of PlantXViT is that it is not able to maintain a lower count of GFLOPs due to the operations involved in the inception block. The approach by \citet{chen2021identifying} has the least GFLOPs. However, it is at the cost of model's classification efficiency on all the parameters. A good balance is needed between the model's memory requirement, computational demands and the desired level of efficiency.  
	
	\begin{table}[!h]
		\centering
		\caption{Comparison of the trainable parameters and FLOPs}
		\label{tab7}
		\scalebox{1}{
			\begin{tabular}{lp{1.5cm}p{1.5cm}cr}
				\hline
				\hline
				\textbf{Approach} & \textbf{Total params(M)} & \textbf{Trainable params(M)} & \textbf{GFLOPs} & \textbf{Memory (MB)} \\
				\hline
				\citet{karthik2020attention} & \textbf{0.72} & \textbf{0.72} & 3.59 & \textbf{2.8} \\
				
				\citet{chen2020attention} & 0.82 & 0.82 & 3.4 & 0.76 \\
				
				\citet{chen2021identifying} & 4.32 & 2.06 & \textbf{0.78} & 16.8 \\
				
				\citet{chen2021identification} & 4.32 & 2.06 & 0.83 & 16.9 \\
				
				\citet{zhao2022ric} & 6.71 & 6.71 & 11.9 & 25.8 \\
				
				\textbf{PlantXViT} & 0.85 & 0.85 & 11.8 & 3.4 \\
				\hline
				\hline
			\end{tabular}	
		}
	\end{table}
	
	\section{Conclusion}
	\label{conclusion}
	In the present work, a ViT enabled CNN model is proposed for plant disease detection and identification. The model combines the benefits of the transformer and CNN in achieving higher precision and accuracy in terms of its feature extraction capability and classification performance. The experimental results demonstrate that the model's performance is impressive on five publicly available datasets of different sizes with images captured under varying background conditions. In the experiment for plant disease detection, PlantXViT achieves 93.55\%, 89.24\%, 92.59\%, 98.86\%, and 98.33\% overall accuracy on Apple, Embrapa, Maize, PlantVillage, and Rice datasets, respectively. It is remarkable to note that the model outperforms other state-of-the-art models in efficiently identifying plant diseases. The model is also evaluated for the interpretability of its prediction results using Grad-CAM and LIME methods and the results show that the model's results are reasonably interpretable. The only hindrance in the applicability of PlantXViT is its higher computational requirement. In future, it is planned to work on reducing the FLOPs count while maintaining the model's efficiency and explainability.
	
	\bibliographystyle{unsrtnat}  
	\bibliography{references}

\end{document}